\RequirePackage[]{graphicx} 

\def\mode{1} 

\if 1\mode

    \documentclass{article}

    \usepackage[a4paper,bindingoffset=0.1in,left=1in,right=1in,top=1in,bottom=1in,footskip=.25in,headsep=.1in]{geometry}

    \usepackage[toc,page]{appendix}
    \usepackage[hidelinks]{hyperref}
    \hypersetup{colorlinks, linkcolor={red},citecolor={green}, urlcolor={blue}}

    \usepackage{fancyhdr}
    \usepackage{titlesec}
    \titleformat{\subsubsection}[runin] 
                {\normalfont\bfseries}  
                {\thesubsubsection}     
                {0.5em}                 
                {}                      
                [:]                     
    \titlespacing*{\subsubsection}{0pt}{0.5em}{0.5em}
    
\fi

\if 2\mode
    \documentclass[submit, journal,twoside,web]{ieeecolor}
\fi

\usepackage[usenames,dvipsnames]{xcolor}
\usepackage{etoolbox}
\newtoggle{arXiv}
\newtoggle{TMI}

\newtoggle{SUPMATERIAL}
\togglefalse{SUPMATERIAL}

\newtoggle{TOC}
\togglefalse{TOC}

\if 1\mode
    \toggletrue{arXiv}
    \togglefalse{TMI}
\fi

\if 2\mode
    \toggletrue{TMI}
    \togglefalse{arXiv}
\fi

\iftoggle{TMI}{
    \usepackage{tmi}
    \usepackage[hidelinks]{hyperref}
    \hypersetup{colorlinks, linkcolor={nblue}, urlcolor={black}}
    \markboth{\journalname, VOL. XX, NO. XX, XXXX 2022}
    {Luo \MakeLowercase{\textit{et al.}}: Preparation of Papers for IEEE TRANSACTIONS and JOURNALS (Jan 2022)}
    \usepackage{amsfonts} 
    \DeclareMathSymbol{\shortminus}{\mathbin}{AMSa}{"39}
}

\usepackage[utf8]{inputenc}
\usepackage{amsmath}
\usepackage{bm}
\usepackage{bbold}
\usepackage{authblk}
\usepackage{amssymb}
\usepackage{graphicx}
\usepackage{ifdraft}
\usepackage{marginnote}
\usepackage{pdfpages}
\usepackage{tabularx,booktabs}
\usepackage{algorithm}
\usepackage{algpseudocode}
\usepackage{adjustbox}
\usepackage{booktabs}
\usepackage{cite}
\usepackage{makecell}
\iftoggle{TMI}{\usepackage[capitalise, nameinlink]{cleveref}}
{\usepackage[capitalise, noabbrev]{cleveref}}
\creflabelformat{equation}{#2\textup{#1}#3}
\usepackage{blindtext}
\usepackage{times}
\usepackage{textcomp}
\usepackage{subcaption}
\usepackage{ragged2e}

\graphicspath{{./figures/}}
\usepackage[skip=1pt, labelsep=period]{caption}
\usepackage{mdframed}
\mdfdefinestyle{mdstyle}{align=center,innertopmargin=4pt,
    innerleftmargin=3pt,innerrightmargin=3pt,innerbottommargin=2pt,backgroundcolor=gray!50,linecolor=gray!50}
\definecolor{gray}{RGB}{220,220,220}

\usepackage{color}

\usepackage{changes}
\newcommand{\stkout}[1]{\ifmmode\text{\sout{\ensuremath{#1}}}\else\sout{#1}\fi}
\setdeletedmarkup{\stkout{#1}}

\title{MRI Reconstruction Using Deep Bayesian {Estimation}}

\newcommand{\authorA}{Guanxiong Luo}
\newcommand{\authorB}{Na Zhao}
\newcommand{\authorC}{Wenhao Jiang}
\newcommand{\authorD}{Edward S. Hui}
\newcommand{\authorE}{Peng Cao}

\newcommand{\affilA}{Department of Diagnostic Radiology, The University of Hong Kong, Hong Kong}
\newcommand{\affilB}{Department of Statistics and Actuarial Science, The University of Hong Kong}

\newcommand{\corAdress}{Peng Cao, Department of Diagnostic Radiology, The University of Hong Kong, Hong Kong,
Address: 5 Sassoon Road, Pok Fu Lum, Hong Kong, China.}

\newcommand{\corMail}{Email: caopeng1@hku.hk}

\author[1]{\authorA}
\author[2]{\authorB}
\author[1]{\authorC}
\author[1]{\authorD}
\author[1]{\authorE, \thanks{\corAdress\,\corMail}}

\affil[1]{\affilA}
\affil[2]{\affilB}

\begin{document}
\maketitle
\begin{abstract}
\noindent
\textbf{Purpose:} To develop a deep learning-based Bayesian estimation for MRI reconstruction. \\
\noindent
\textbf{Methods:} We modeled the MRI reconstruction problem with Bayes's theorem, following the recently proposed PixelCNN++ method. The image reconstruction from incomplete k-space measurement was obtained by maximizing the posterior possibility.
A generative network was utilized as the image prior, which was computationally tractable, and the k-space data fidelity was enforced by using an equality constraint. The stochastic backpropagation was utilized to calculate the descent gradient in the process of maximum a posterior, and a projected subgradient method was used to impose the equality constraint. In contrast to the other deep learning reconstruction methods, the proposed one used the likelihood of prior as the training loss and the objective function in reconstruction to improve the image quality.\\
\noindent
\textbf{Results:} The proposed method showed an improved performance in preserving image details and reducing aliasing artifacts, compared with GRAPPA, $\ell_1$-ESPRiT, model-based deep learning architecture for inverse problems (MODL), and {variational network (VN)}, last two were state-of-the-art deep learning reconstruction methods. The proposed method generally achieved more than {3} dB peak signal-to-noise ratio improvement for compressed sensing and parallel imaging reconstructions compared with the other methods.\\
\noindent
\textbf{Conclusion:} The Bayesian estimation significantly improved the reconstruction performance, compared with the conventional $\ell_1$-sparsity prior in compressed sensing reconstruction tasks.  More importantly, the proposed reconstruction framework can be generalized for most MRI reconstruction scenarios.\\
\end{abstract}

\section{Introduction}
In compressed sensing MRI reconstruction, the commonly used analytical regularization such as $\ell_1$ regularization can ensure the convergence of the iterative algorithm and improve MR image quality \cite{lustig2007sparse}. The conventional iterative reconstruction algorithm with analytical regularization has an explicit mathematical deduction in gradient descent, which ensures the convergence of the algorithm to a local or global optimal and the generalizability. Besides, the dictionary learning is an extension of analytical regularization, providing an improvement over the $\ell_1$ regularization in specific applications \cite{ravishankar2010mr}.
The study of analytically regularized reconstruction mainly focused on choosing the appropriate regularization function and parameters. As an extension of analytical regularization, the deep learning reconstruction was employed as an unrolled iterative algorithm for solving the regularized optimization \cite{sun2016deep,aggarwal2018model} or used as a substitute for analytic regularization \cite{schlemper2017deep,mardani2018deep}.
With the advances of deep learning methodology, research started shifting the paradigm to structured feature representation {and recovery} of MRI, such as cascade \cite{sun2016deep}, {iterative with fidelity term} \cite{schlemper2017deep,mardani2018deep,aggarwal2018model}, and generative \cite{mardani2018deep} deep neural networks. 
Especially, the method proposed in \cite{sun2016deep} recast the compressed sensing reconstruction into a specially designed neural network that still partly imitated the analytical data fidelity and regularization terms. In that study, the analytical regularization term was replaced with convolutional layers and a specially designed activation function \cite{sun2016deep}. 
{In a latter study, several data-consistency layers were embedded in a feed-forward convolutional network to keep the reconstructed image consistent with k-space data \cite{schlemper2017deep}. In another study, a discriminator (from a generative adversarial network) was used to create an image manifold, ensuring the reconstruction can explore the feasible data space, along with the data consistency layers \cite{mardani2018deep}. In a recent study, following the well-known "unrolling" of iterative reconstruction approach, a semi-iterative convolutional network was also developed as a model-based deep learning architecture for inverse problems (MODL) \cite{aggarwal2018model}. In a more recent study, a sophisticated framework for generalized compressed sensing reconstruction was formulated as a variational network (VN) that was embedded in an unrolled gradient descent scheme \cite{hammernik2018learning}.}
These deep learning methods may show improved performance in some predetermined acquisition settings or pre-trained imaging tasks. However, they also lack flexibility when used with changes in MRI under-sampling scheme, the number of radio-frequency coils, and matrix size or spatial resolution. Such restriction is caused by the embedment of k-space data fidelity and the regularization terms into neural network implementations. Therefore, it was preferable to separate the k-space data fidelity and neural network-based regularization for improving the flexibility in changing MRI acquisition configurations. 

This study applied Bayesian estimation to model the MRI reconstruction problem, and the statistical representation of an MRI database was used as a prior model. In Bayesian estimation, the prior model is required to be computationally scalable and tractable \cite{rezende2014stochastic,arridge2019solving}. The scalability {indicates that the prior model has an explicit  probabilistic distribution function, which can be used as loss function for both network training and image reconstruction} \cite{rezende2014stochastic,arridge2019solving,goodfellow2016}. The tractability of the prior model means the probability of an given image can be calculated directly without any approximation, besides the gradient that facilitates the maximization of posterior distribution can be calculated by stochastic backpropagation\cite{rezende2014stochastic,arridge2019solving}.
In such Bayesian estimation, the image to be reconstructed was referred to as the parameters of the Bayesian model, which was conditioned on the measured k-space data (as the posterior).
Bayes's theorem expressed the posterior as a function of the k-space data likelihood and the image prior. {Recently, a variational autoencoder (VAE) was applied as the deep density prior in MRI reconstruction, under a Bayesian estimation framework  \cite{tezcan2018mr}, but the evidence lower bound (ELBO) used in VAE is an approximation density model. The ELBO is calculated via Monte Carlo sampling which leads to expensive computational cost. Moreover, the general challenge of VAE is to match the conditional distribution in latent space to the explicit distribution\cite{rosca2018}.}
For the image prior, Refs \cite{salimans2017pixelcnn, oord2016pixel} proposed a generative deep learning model, providing a tractable and scalable likelihood. In those studies, the image prior model was written as the multiplication of the conditional probabilities those indicated pixel-wise dependencies of the input image.
The k-space data likelihood described how the measured k-space data was computed from a given MR image. The relationship between k-space data and MR image can be described, using the well-known MRI encoding matrix in an equality constraint \cite{fessler2010model}.
With such computationally scalable and tractable prior model, the maximum a posterior can serve as an effective estimator \cite{arridge2019solving} for the high dimensional image reconstruction problem tackled in this study.
To summarize, the Bayesian estimation for MRI reconstruction had two separate models: the k-space likelihood model that was used to encourage data consistency and the image prior model that was used to exploit knowledge learned from an MRI database.

This paper presented a generic and interpretable deep learning-based reconstruction framework, using Bayesian estimation. It employed a generative network as the MR image prior model. The proposed framework was capable of exploiting the MR image database with the prior model, regardless of the changes in MR imaging acquisition settings. Also, the reconstruction was achieved by a series of estimations those employed the maximum likelihood of posterior with the image prior, i.e., applying the Bayesian estimation repeatedly. The reconstruction iterated over the data fidelity enforcement in k-space and the image refinement, using the Bayesian estimation. During the iteration, the projected sub-gradient algorithm was used to maximize the posterior. The method is theoretically described, which was adapted from the methodology proposed by others \cite{salimans2017pixelcnn}, and then demonstrated in different MRI acquisition scenarios, including parallel imaging, compressed sensing, and non-Cartesian reconstructions. The robustness and the reproducibility of the algorithm were also experimentally validated.
\section{Theory}
The proposed method applied a generative neural network, as a data-driven MRI prior, to an MRI reconstruction method. This section contained an MRI reconstruction method using Bayes$'$ theorem and a generative neural network-based MRI prior model, a pixel-wise joint probability distribution for images, using the PixelCNN++ \cite{salimans2017pixelcnn}. 
\subsection{MRI reconstruction using Bayes$'$ theorem}
With Bayes$'$ theorem, one could write the posterior as a product of likelihood and prior:
\begin{equation}
f(\boldsymbol{x}|\boldsymbol{y}) = \frac{f(\boldsymbol{y}\mid\boldsymbol{x})g(\boldsymbol{x})}{f(\boldsymbol{y})}    \propto f(\boldsymbol{y}\mid \boldsymbol{x} )\,g(\boldsymbol{x} )
\label{eq:1}
\end{equation}
where $f(\boldsymbol{y}\mid\boldsymbol{x})$ is probability of the measured k-space data {$ \boldsymbol{y}\in \mathbb{C}^M$} for a given image {$\boldsymbol{x}\in \mathbb{C}^N$}, { $N$ is the number of pixels and $M$ is the number of measured data points}. $g(\boldsymbol{x})$ is the prior model {that estimates the distribution of MR images}. 
{In order to avoid the confusion with the likelihood occurs in modelling the prior, $f(\boldsymbol{y}\mid\boldsymbol{x})$ is referred to as k-space likelihood model.}
The image reconstruction is achieved by exploring the posterior $f(\boldsymbol{x}\mid\boldsymbol{y})$ with an appropriate estimator. The maximum a posterior estimation (MAP) could provide the reconstructed image $\hat{\boldsymbol{x}}$ that is given by:
\begin{equation}
{\hat {\boldsymbol{x} }}_{\mathrm {MAP} }(\boldsymbol{y})={\underset {\boldsymbol{x}}{\operatorname {arg\,max} }}\ f(\boldsymbol{x} \mid \boldsymbol{y})={\underset {\boldsymbol{x}}{\operatorname {arg\,max} }}\ f(\boldsymbol{y}\mid \boldsymbol{x})\,g(\boldsymbol{x} )
\label{eq:2}
\end{equation}
{In this way, the reconstruction problem is recast as posterior probability calculations.}
{\cref{eq:2} indicates two models are required for the estimation: the prior model, $g(\boldsymbol{x})$, and k-space likelihood model, $f(\boldsymbol{y}\mid\boldsymbol{x})$.}
\begin{figure}[H]
    \begin{mdframed}[backgroundcolor=gray!50,linecolor=gray!50]
	\centering
	\includegraphics[width=\textwidth]{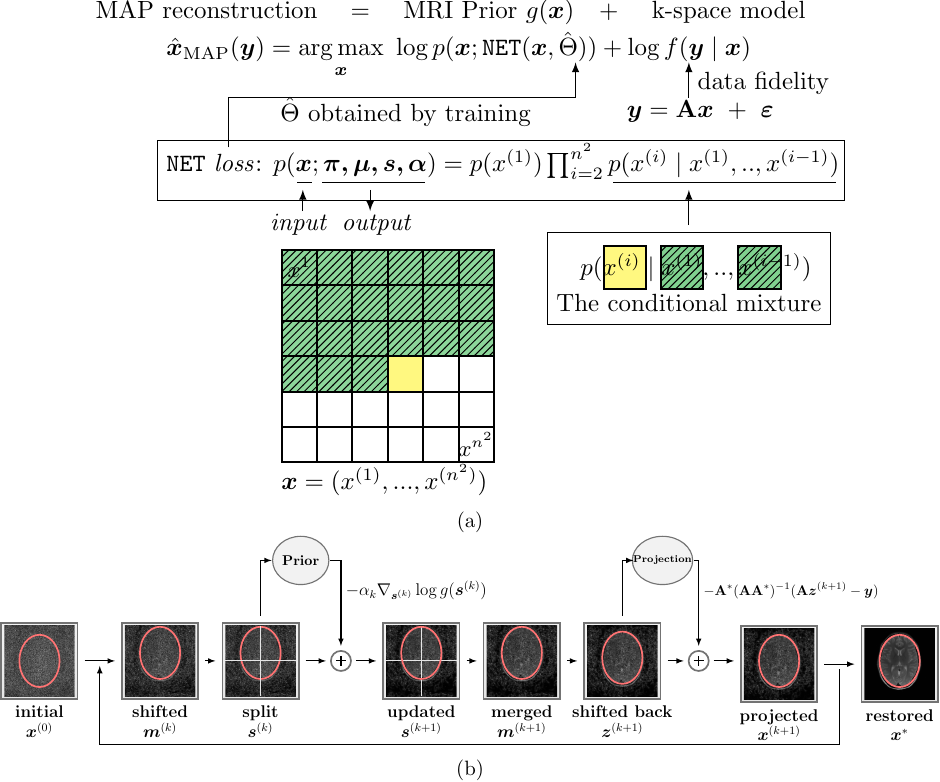}
    \end{mdframed}
	\caption{(\textbf{a}) Overview of the MAP reconstruction. Conditional model in \cite{salimans2017pixelcnn,oord2016pixel} defined the probability of image pixel (yellow) $x_{i,j}$ dependent on all the pixels from its up and left side (green). (\textbf{b}) In this method, we reconstructed images with $256\times256$ matrix size, using the prior model $g(\boldsymbol{x})$ that was trained with $128\times128$ images and illustrated in Supporting Figure S1. To reconcile this mismatch, we split one $256\times256$ image into four $128\times128$ patches for applying the prior model. After updating $\boldsymbol{s}^{(k+1)}$, four patches for one image were merged to form an image with the original size of $256\times256$. Then the merged image was projected onto $\{\boldsymbol{x}\mid\boldsymbol{y} = \mathbf{A} \boldsymbol{x} \ + \ \boldsymbol{\varepsilon}\}$ in \cref{eq:13}. Furthermore, the random shift along phase encoding direction was applied to mitigate the stitching line in-between patches.
	}
	\label{fig:fig_combinednet}
\end{figure}
\subsection{{Expressing prior using model parameters}}
{The proposed method estimated the distribution of MR image features concerning parameters of a mixture distribution model given by a generative network \cite{salimans2017pixelcnn}. The mixture distribution model (or the distribution of MR image features) was then used to compute the likelihood of images, serving as a prior model for reconstruction (as illustrated in \cref{fig:fig_combinednet}). The generative network was commonly used as a parameterized model, approximating the distribution of image features \cite{salimans2017pixelcnn}. There were three reasons for using such mixture distribution model and generative network approximation \cite{salimans2017pixelcnn}: 1) hierarchical architectures allowed the network to capture complex structure in the image, {2}) the mixture of conditional distribution characterized an intrinsic dependence among pixels and factorized the probability density function for an image over its pixels, {3}) they discretized/instanced logistic mixture likelihood for using pixelCNN++} \cite{salimans2017pixelcnn}.
For MRI reconstruction, {the number of image channels was changed from three (i.e., RGB channels for color image) to two (i.e., real and imaginary parts for MR image).} For each image pixel, {the continuous variable $\nu$ denoted the real or the imaginary part of a pixel, $\nu=\operatorname{Re}(x_{i,j})\ \text{or}\ \operatorname{Im}(x_{i,j})$}. Like in the VAE and pixelCNN++  \cite{salimans2017pixelcnn,kingma1606improving}, the distribution of $\nu$ was a mixture of the logistic distribution, given by
\begin{equation}
    \nu \sim \sum_{k=1}^{K} \pi_k \mathrm{logistic}(\mu_k, s_k).
    \label{eq:3}
\end{equation}
Here, $\pi_i$ was the mixture indicator, $\mu_i$ and $s_i$ were the mean and scale of logistic distribution, $K$ was the number of mixture distributions, respectively. {It should be noted that the logistic distribution function was interchangeable with any distribution function for creating a mixture with sufficient representation capacity. However, for a tractable computation, the logistic distribution was recommended in the previous study \cite{salimans2017pixelcnn}.} Then the probability on each observed $\nu$ of the pixel was computed as \cite{salimans2017pixelcnn}
\begin{equation}
    P(\nu;\pi, \mu, s) = \sum_{k=1}^{K}\pi_k[\sigma((\nu+d/2-\mu_k)/s_k)-\sigma((\nu-d/2-\mu_k)/s_k)],
    \label{eq:4}
\end{equation}
where $\sigma$ was the logistic sigmoid function, {$d$ was the smallest discretized interval  for $\nu$.}
Furthermore, in \cite{salimans2017pixelcnn,oord2016pixel}, each pixel was dependent on all previous pixels up and to the left in an image, as shown in \cref{fig:fig_combinednet}. {The derivation from \cref{eq:3} to \cref{eq:4} is provided in Appendix A.}
The conditional distribution of the subsequent pixel $(\operatorname{Re}(x_{i,j}), \operatorname{Im}(x_{i,j}))$ at position $(i,j)$ was given by \cite{salimans2017pixelcnn}
\begin{align*}
p(x_{i,j}|C_{i,j}) = p(\operatorname{Re}(x_{i,j}), \operatorname{Im}(x_{i,j})\mid C_{i,j}) =& P(\operatorname{Re}(x_{i,j});\pi{(C_{i,j})},\mu_{\operatorname{Re}}(C_{i,j}), s_{\operatorname{Re}}(C_{ij}))\times\\ &P(\operatorname{Im}(x_{i,j});\pi{(C_{i,j})},\hat{\mu}_{\operatorname{Im}}(C_{i,j}, \operatorname{Re}(x_{i,j})),s_{\operatorname{Im}}(C_{i,j}))\nonumber\\
    \hat{\mu}_{\operatorname{Im}}(C_{i,j},\operatorname{Re}(x_{i,j}))&= \mu_{\operatorname{Im}}(C_{i,j}) + \alpha(C_{i,j})\operatorname{Re}(x_{i,j})\ ,
\end{align*}
\noindent
where the $C_{i,j}=\{x_{i-1,j},x_{i-2,j},...,x_{1,1}\}$ denoted the context information which was comprised of the mixture indicator and the previous pixels as showed in \cref{fig:fig_combinednet}, $\alpha$ was the coefficient related to mixture indicator and previous pixels, {$\mu_{\operatorname{Re}}$ was the $\mu$ of the real channel of $C_{i,j}$, and  $\mu_{\operatorname{Im}}$, $s_{\operatorname{Re}}$,  $s_{\operatorname{Im}}$ were subscripted in the same way.} $p(\operatorname{Re}(x_{i,j}), \operatorname{Im}(x_{i,j})\mid C_{i,j})$ was also a joint distribution for both real and imaginary channels. The real part of the first pixel, i.e., $x^{(1)}=x_{1,1}$ in \cref{fig:fig_combinednet}, was predicted by a mixture of logistics as described in \cref{eq:3}. This definition assumed that the mean of mixture components of the imaginary channel was linearly dependent on the real channel.  In this study, the number of mixture components was 10.
In this model, mixture indicator was shared between two channels. The $n\times n$ image could be considered as an vectorized image $\mathit{\boldsymbol{x}} = (x^{(1)},...,x^{(n^2)})$ by stacking pixels from left to right and up to bottom of one another, i.e., $x^{(1)}=x_{1,1}, x^{(2)}=x_{2,1},...,$ and $x^{(n^2)}=x_{n,n}$ {, as illustrated in \cref{fig:fig_combinednet}}. The joint distribution of the image vector could be expressed as following \cite{salimans2017pixelcnn}:
\begin{equation}
    p(\boldsymbol{x};\boldsymbol{\pi,\mu,s, \alpha})=p(x^{(1)})\prod_{i=2}^{n^2} p(x^{(i)}\mid x^{(1)},..,x^{(i-1)}).
    \label{eq:p_model}
\end{equation}
$\boldsymbol{\pi,\mu,s,\alpha}$ were the parameters of mixture distribution for each pixel {as were defined in Equations. 4-6}. The generative network PixelCNN++ was expected to predict the joint probability distribution of all pixels in the input image \cite{salimans2017pixelcnn}. Therefore, the network $\mathtt{NET}(\boldsymbol{x}, \Theta)$ was trained by maximizing the likelihood in \cref{eq:p_model}, as the training loss was given by
\begin{equation}
  \hat{\Theta} = {\underset {\Theta }{\operatorname {arg\,max}\ }}{ p(\boldsymbol{x}; \mathtt{NET}(\boldsymbol{x}, \Theta))},
\end{equation}
where $\Theta$ contained the trainable {parameters} within the network. After training, the network could be used as the image prior. Here, we defined the prior model $g(\boldsymbol{x})$ as
\begin{equation}
    g(\boldsymbol{x}) = p(\boldsymbol{x}; \mathtt{NET}(\boldsymbol{x}, \hat{\Theta})).
    \label{eq:g_prior}
\end{equation}
{For example, if the complex images $\boldsymbol{x}$ for training had a size of (128,128), and the number of mixture distribution $K = 10$, then $\boldsymbol{\pi,\mu,s,\alpha}$ would get dimensions of (128,128,10,1), (128,128,10,2), (128,128,10,2), and (128,128,10,1), respectively.}
To summarize, a prior model of $\boldsymbol{x}$ was defined in Eqs. from \ref{eq:2} to \ref{eq:g_prior} that could be considered as a data-driven model, utilizing the knowledge learned from an image database.
\subsection{Image reconstruction by MAP}
  The measured  k-space data $\boldsymbol{y}$ was given by
  \begin{equation}
      \boldsymbol{y} = \mathbf{A} \boldsymbol{x} \ + \ \boldsymbol{\varepsilon},
      \label{eq:mri_ax_b}
  \end{equation}
where $\mathbf{A}$ was the encoding matrix, $\boldsymbol{x}$ was MR image, and $\boldsymbol{\varepsilon}$ was the noise. The matrix {$\mathbf{A}=GFS$}, {where $S:\mathbb{C^N}\rightarrow\mathbb{C^{N\times\gamma}}$ coil sensitivity maps, $\gamma$ the number of coils, $F:\mathbb{C}^{N\times\gamma}\rightarrow\mathbb{C}^{N\times\gamma}$ the Fourier operator, and $G:\mathbb{C}^{N\times\gamma}\rightarrow\mathbb{C}^{M\times\gamma}$ the k-space sampling operator/mask.}
{In this study, the additive noise $\varepsilon$ was assumed to follow a Gaussian distribution with zero mean.}
Substituting \cref{eq:g_prior} into the log-likelihood for \cref{eq:2}  yielded
\begin{equation}
    \hat{\boldsymbol{x}}_\mathrm{MAP}(\boldsymbol{y}) =\underset {\boldsymbol{x}}{\operatorname {arg\,max} }\ \log{f(\boldsymbol{y}\mid \boldsymbol{x} )} + \log{p(\boldsymbol{x}; \mathtt{NET}(\boldsymbol{x}, \hat{\Theta}))}.
    \label{eq:9}
\end{equation}
From the data model, the log-likelihood term for ${f(\boldsymbol{y}\mid \boldsymbol{x} )}$ had less uncertainty, considering the MR imaging principles, for a given image $\boldsymbol{x}$, the probability for k-space, $\boldsymbol{y}$, i.e., ${f(\boldsymbol{y}\mid \boldsymbol{x} )}$ when $\boldsymbol{y} = \mathbf{A} \boldsymbol{x}+ \ \boldsymbol{\varepsilon}$, was close to a constant with the uncertainty from noise that was irrelevant and additive to $\boldsymbol{x}$ . Hence, \cref{eq:9} could be rewritten as 
\begin{equation}
    {\hat {\boldsymbol{x} }}_{\mathrm {MAP} }(\boldsymbol{y}) =  {\underset {\boldsymbol{x}}{\operatorname {arg\,max} }} \, \log{p(\boldsymbol{x}; \mathtt{NET}(\boldsymbol{x}, \hat{\Theta}))} \qquad \mathrm{s.t.} \quad \boldsymbol{y} = \mathbf{A} \boldsymbol{x} \  + \ \boldsymbol{\varepsilon}
    \label{eq:12}
\end{equation}
The equality constraint for data consistency was the result of eliminating the first log-likelihood term in \cref{eq:9}. The projected subgradient method was used to solve the equality constrained problem \cite{gregor2013deep,boyd2003subgradient}. In \cite{boyd2003subgradient}, authors proposed a stochastic backpropagation method for computing gradients through random variables for deep generative models. In PixelCNN++, the stochastic backpropagation provided the subgradient $\nabla_{\boldsymbol{x}}\log{g(\boldsymbol{x})}$, where $g(\boldsymbol{x}) = p(\boldsymbol{x}; \mathtt{NET}(\boldsymbol{x}, \hat{\Theta}))$, for minimizing the log-likelihood in \cref{eq:12}. We empirically found that the dropout (which applied to {gradient update}) was necessary, when using the gradient to update $\boldsymbol{x}$ in \cref{eq:12} \cite{pmlr-v48-gal16}. To summarize, the MAP-based MRI reconstruction had the following iterative steps:
 \begin{description}
 	\item {\textbf{Repeat}}
    \item \quad Get the descent direction $\nabla_{\boldsymbol{x}^{(k)}}\log{g(\boldsymbol{x}^{(k)})}$
    \item \quad Pick up a step size $\alpha_k=1/k$ or {use a fixed step size}
    \item \quad Update $\boldsymbol{z}^{(k+1)}=\boldsymbol{x}^{(k)}-\alpha_k \nabla_{\boldsymbol{x}^{(k)}}\log{g(\boldsymbol{x}^{(k)})}$
    \item \quad Projection $\boldsymbol{x}^{(k+1)}={\underset {\boldsymbol{x}\in X }{\operatorname {arg\,min} }}\frac{1}{2}\|\boldsymbol{x}-\boldsymbol{z}^{(k+1)}\|_2^2$
    \item {\textbf{Until} $\|\mathbf{A}\boldsymbol{z}-\boldsymbol{y}\|_2^2 < \epsilon$ or $k > \textrm{maxIter}$}
 \end{description}

 The projection of $\boldsymbol{z}$ onto  $\{\boldsymbol{x}\mid\boldsymbol{y} = \mathbf{A} \boldsymbol{x} \ + \ \boldsymbol{\varepsilon}\}$ was given by
\begin{equation}
    \mathcal{P}(\boldsymbol{z}) = \boldsymbol{z} - \mathbf{A}^*(\mathbf{A}\mathbf{A}^*)^{-1}(\mathbf{A}\boldsymbol{z}-\boldsymbol{y}).
    \label{eq:13}
\end{equation}
Therefore, the generative network as a prior model was incorporated into the reconstruction of $\boldsymbol{x}$ through the Bayesian inference based on MAP.

%
\section{Methods}
\subsection{MRI data and pre-processing} 
Both knee and brain MRI data were used to test the reconstruction performance 
 of the proposed method. The knee MRI data (multi-channel k-space data, 973 scans) were downloaded from fastMRI reconstruction database \cite{DBLP:journals/corr/abs-1811-08839}. As such, NYU fastMRI investigators provided data but did not participate in analysis or writing of this report. A listing of NYU fastMRI investigators, subject to updates, can be found at: fastmri.med.nyu.edu. The primary goal of fastMRI is to test whether machine learning can aid in the reconstruction of medical images.
The knee data had two contrast weightings: proton-density with and without fat suppression (PDFS and PD). Scan parameters included 15-channel knee coil and 2D multi-slice turbo spin-echo (TSE) acquisition, and other settings which could be found in Ref. \cite{DBLP:journals/corr/abs-1811-08839}. 


For brain MRI, we collected 2D multi-slice T1 weighted, T2 weighted, T2 weighted {fluid-attenuated inversion recovery} (FLAIR), and T2$^*$ weighted brain images from 16 healthy volunteers examined with clinical standard-of-care protocols, approved by the Institutional Review Board of The University of Hong Kong/Hospital Authority Hong Kong West Cluster. The consent was obtained from all volunteers. All brain data were acquired using our 3T MR scanner (Philips, Achieva), and an eight-channel brain RF coil. T1 weighted, T2 weighted, and T2 weighted FIAIR images were all acquired with TSE readout. Meanwhile, T2$^*$-weighted images were obtained using a gradient-echo sequence. Brain MRI parameters for four contrast weightings were listed in Supporting Information Table S1.

Training images were reconstructed from multi-channel k-space data without undersampling. Then, these image datasets after coil combination were scaled to a magnitude range of $[-1,1]$ and resized to an image size of 256$\times$256. The training of PixelCNN++ model required a considerable computational capacity when a large image size was used. In this study, the $128\times128$  was the largest size that our 4-{graphics processing unit} (GPU) server could handle. Hence, the original $256\times256$ images were resized into $128\times128$ low-resolution images by cropping in k-space for knee MRI. For brain MRI, we split each raw  $256\times256$ image into four $128\times128$ image patches, before fed into the network for training.
Real and imaginary parts of all 2D images were separated into two channels when inputted into the neural network.
{For the data partitioning, we first separated all multi-slice volumes into training and testing groups. Then we split the volume into slices (i.e., 2D images).}
For knee MRI, 15541 images were used as the training dataset, and 170 images were used for testing. For brain MRI, 1300 images were used as the training dataset, and 300 images were used for testing. {For data analysis, root-mean-square error (RMSE in \%), peak signal-to-noise ratio (PSNR, in dB) and structural similarity index (SSIM, in \%) were used to quantify the image accuracy.} 
\subsection{Deep neural network}
{
As illustrated in \cref{fig:fig_combinednet}, when predicting the current pixel, i.e., yellow square in \cref{fig:fig_combinednet} or $x_{i,j}$, pixels to the left and above were used as the inputs for the estimation, i.e., green squares or $C_{i,j}=\{x_{i-1,j},x_{i-2,j},...,x_{1,1}\}$. In \cite{van2016conditional}, the convolution stream was split into two network stacks: one conditioned on the ``current row so far", i.e., $\{x_{i-1,j},x_{i-2,j},...,x_{1,j}\}$, and another conditioned on all rows above, i.e., $\{x_{n,j-1},x_{n-1,j-1},...,x_{1,1}\}$, creating up-stream and down-stream as shown in Supporting Information Figure S1. The upstream and downstream first pass through two blocks that had ``padding" and ``shifting" functions for removing blinding spots \cite{van2016conditional}. Then paired residual blocks (Res-blocks) that contained three gated ResNets were applied. A gated ResNet had three convolutional layers with the middle layer as a gated layer \cite{van2016conditional}. In the Res-block, each convolution layer had 100 filters with a kernel size of 3$\times$3. In between the first and second Res-blocks, as well as the second and third Res-blocks, the network contained subsampling operations, implemented using a 2$\times$2 stride convolution. In between the fourth and fifth Res-blocks, as well as the fifth and sixth Res-blocks, the network had a transpose stride convolution, i.e., 2$\times$2-upsampling. These subsampling and upsampling caused information loss; therefore, the network also employed short-cut connections in-between Res-blocks to recover information. The short-cut connections went from the layers in the first Res-block to the corresponding layers in the sixth Res-block, and similarly between Res-blocks two and five, and Res-blocks three and four.}\par
The PixelCNN++ was modified from the code in https://github.com/openai/pixel-cnn. We implemented the reconstruction algorithm using Python, as explained in \cref{eq:13} and Appendix. 
With the trained prior model, we implemented the iterative reconstruction algorithm for maximizing the posterior while enforcing the k-space data fidelity (as explained in Appendix and \cref{fig:fig_combinednet}.
Only two deep learning models were trained and utilized, one for knee MRI with two contrast weightings, and another for brain MRI with four contrast weightings. These two models can support all experiments performed in this study with variable undersampling patterns, coil sensitivity maps, channel numbers, image sizes, and trajectory types.  Our networks were trained in Tensorflow software, and on four NVIDIA RTX-2080Ti graphic cards. Other {parameters} were 500 epochs, batch size = 4, and Adam optimizer. It took about four days to train the network for knee dataset and two days for brain dataset under the above-mentioned configuration.

\subsection{Parallel imaging and $\ell_1$ or $\ell_2$ regularization reconstruction}
The {generalized autocalibrating partial parallel acquisition} (GRAPPA) reconstruction was performed with a block size of 4 and 20 central k-space lines as the auto-calibration area  \cite{griswold2002generalized}. We simulated GRAPPA accelerations with undersampling factors from 2 to 4. The representative undersampling masks were shown in Supporting Information Figure S2.
We chose $l_1$-{eigenvalue approach to autocalibrating parallel MRI} (ESPIRiT) \cite{lustig2007sparse,uecker2014espirit,uecker2015berkeley}, MODL \cite{aggarwal2018model}, {and VN} \cite{hammernik2018learning} as baseline methods for comparison. They were originated from analytical regularizations methods. The $\ell_1$-ESPIRIT exploited the sparsity of image, and the MODL was a deep learning method for compressed sensing reconstruction, trained via minimizing $\ell_2$ or $\ell_1$ reconstruction error. In the $\ell_1$-ESPIRiT reconstruction, we set the  $\ell_1$ regularization parameter to be 0.01.
For the training of MODL, the setting followed Ref \cite{aggarwal2018model} when training MODL to reconstruct the undersampled knee data. The only difference was the k-space mask in Ref \cite{aggarwal2018model} was 2D undersampled, while in the current study, the 1D undersampling was applied. The central 20 k-space lines were sampled which account for 7\% of the full k-space of one $256\times256$ image. The others in the outer region were picked randomly with certain undersampling rate for each single slice. {VN method required fixing the sampling mask \cite{hammernik2018learning}. When training VN, we repeated the process multiple times for using different undersampling masks in this study. The sensitivity maps feed into our method, MODL, and VN were estimated from the central $20\times20$ k-space region, using the ESPIRiT function in Berkeley advanced reconstruction toolbox (BART) \cite{uecker2015berkeley}.}\par 
For the proposed method, MR images with $256\times256$ matrix size were reconstructed, using the prior model in \cref{eq:g_prior} that was trained by $128\times128$ images or image patches. During inference, the $256\times256$ image was split into four $128\times128$ patches for applying the prior model, as shown in \cref{fig:fig_combinednet}. After updating $\boldsymbol{s}^{(k+1)}$, four patches for one image were concatenated to form an image with the original size of $256\times256$, before it was projected onto $\{\boldsymbol{x}\mid\boldsymbol{y} = \mathbf{A} \boldsymbol{x} \ + \ \boldsymbol{\varepsilon}\}$ in \cref{eq:13}. The detailed algorithm was presented in the Appendix {B}. When using the patch-based approach, the random shift along the phase encoding direction was applied to eliminate stitching lines. The random shift would move the discontinuity from edge to the center of patches, which was then erased by the next gradient updates. In Supporting Information Video S1, the evolution of restored image was presented.

\subsection{Non-Cartesian k-space acquisition}
In this experiment, spiral sampled k-space from the acquired T2$^*$-weighted k-space data was simulated. {For spiral k-space sampling, the ground truth came from Cartesian sampling.} The method proposed in Ref \cite{lustig2008fast} was used to design the spiral trajectory. The full k-space coverage required 24 spiral interleaves for the spatial resolution used in this study. Besides, the implementation of non-uniform fast Fourier transform was based on the method in Ref \cite{fessler2003nonuniform}. For comparison, we used the conjugate gradient {sensitivity encoding} (CG SENSE), proposed in Ref \cite{pruessmann2001advances}, as a baseline method. {Our implementation exactly followed the original paper of CG SENSE, which applied sensitivity encoding spatially in an L2 minimization, i.e., $min ||Ax-y||_2$, where $A$ the sensitivity encoding and Fourier transform, $x$ the image and $y$ the undersampled k-space, CG was used as the solver.  
In the iterative CG reconstruction, forward and backward Fourier transforms were performed with NUFFT. The NUFFT function was compiled using the code from Jeffrey A. Fessler’s Lab (https://web.eecs.umich.edu/~fessler/), which used min-max interpolation. }
{
\subsection{Prospective experiment in vivo}
To further validate the feasibility of proposed method, we implemented the prospective k-space undersampling for GRAPPA and compressed sensing in a rapid gradient-echo sequence with TE/TR=16/770 ms. The image size was $256\times256$, and the resolution was $0.9 \times 0.9$ $\textrm{mm}^2$. For GRAPPA, the acquisition was accelerated by a factor of 3, and 20 center lines were sampled. For compressed sensing,  only 15\% phase-encoding lines were acquired, and the 20 center lines were kept. Then, acquired k-space data were normalized by the maximum magnitude of zero-filled and reconstructed images. Finally, the images reconstructed via GRAPPA, $\ell_1$-ESPIRiT, our method, and VN were compared.
}
\section{Results}
\subsection{Parallel imaging}
\begin{figure}[H]
	\centering
	\includegraphics[width=\textwidth]{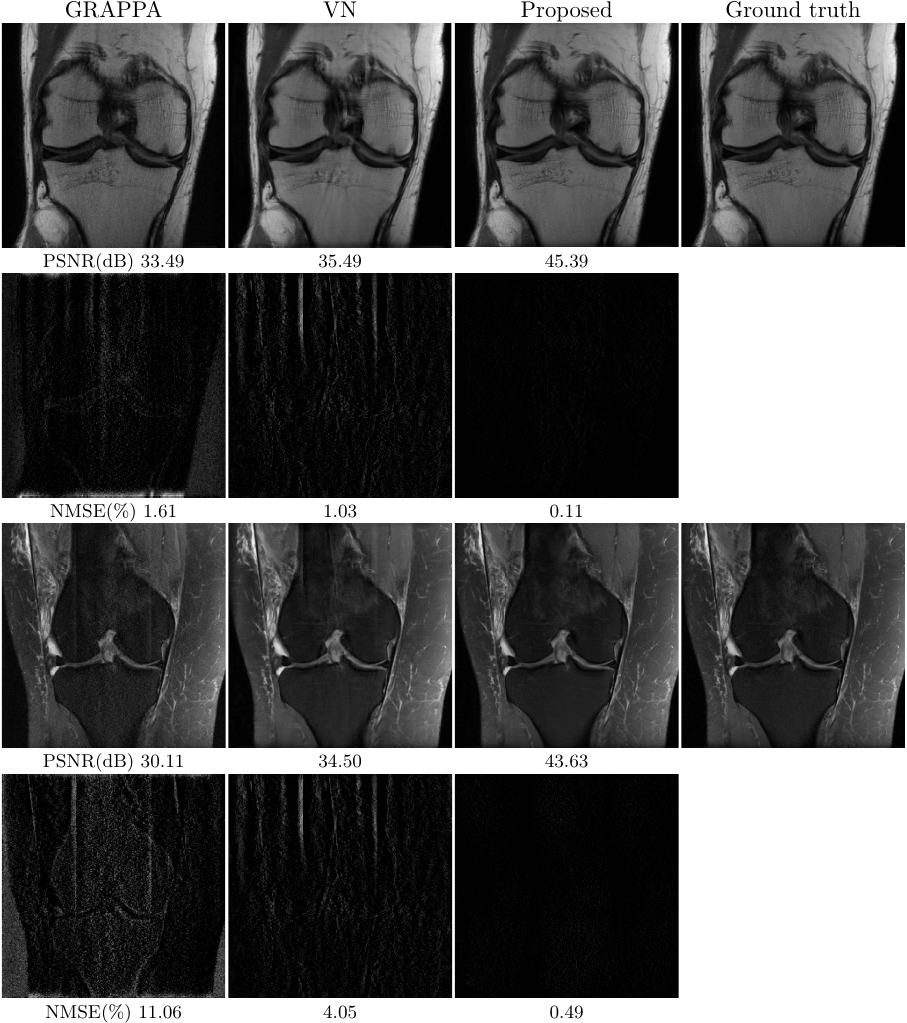}
	\caption{Comparisons on PD and PDFS contrasts using GRAPPA, {VN}, and the proposed reconstructions with R=3 acceleration and $256\times256$ matrix size. The intensity of error maps was five times magnified. The proposed method effectively eliminated noise amplification and aliasing artifact in GRAPPA reconstruction. }
	\label{fig:pi_knee}
\end{figure}

\begin{figure}[H]
	\centering
	\includegraphics[width=0.75\textwidth]{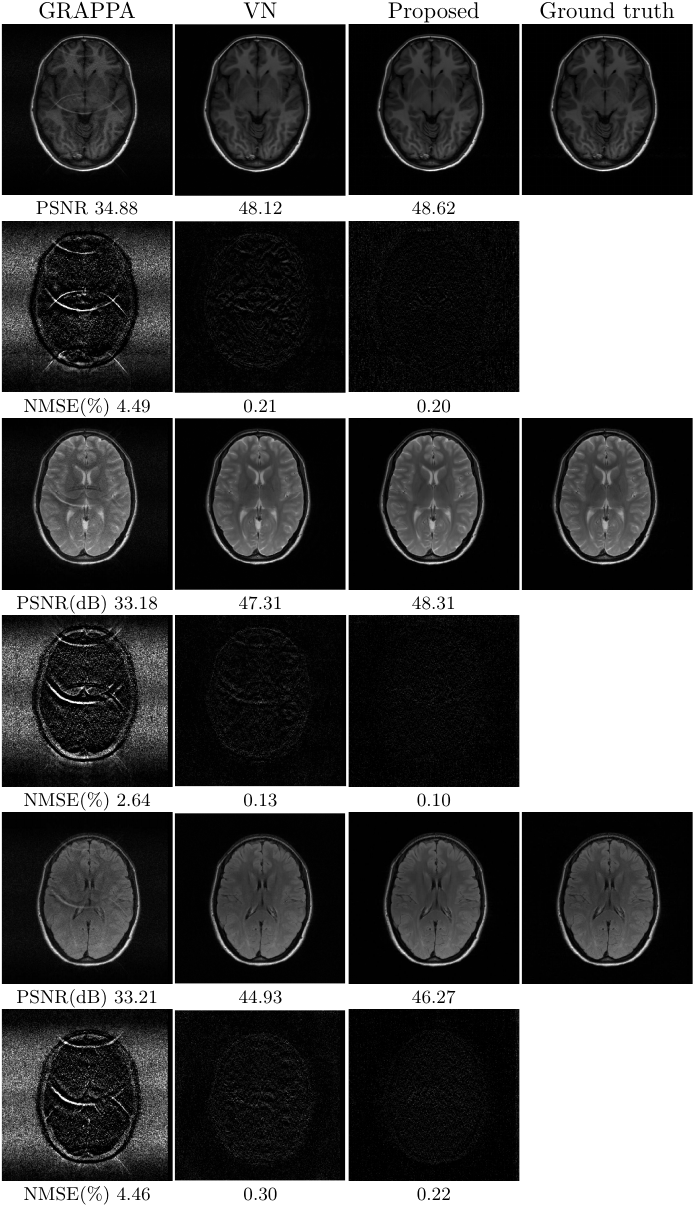}
	\caption{Comparisons on T1, T2, and FLAIR-T2 weighted image reconstruction, using parallel imaging, {VN}, and the proposed reconstruction with R=3 acceleration and $256\times256$ matrix size. The intensity of error maps was 15 times magnified. The proposed method effectively eliminated the noise amplification in GRAPPA reconstruction. }
	\label{fig:pi_brain}
\end{figure}

\cref{fig:pi_knee} and \cref{fig:pi_brain} show the comparison of knee and brain MRI reconstructed using GRAPPA, VN and the proposed method.
The proposed method had an improved performance in recovering brain and knee image details and reducing the aliasing artifacts, compared with GRAPPA {and VN}. As expected, parallel imaging amplified the noise in the low coil sensitivity regions and along the undersampled dimension. On the other hand, error maps demonstrated in \cref{fig:pi_knee} and \cref{fig:pi_brain} showed that both the proposed method {and VN} effectively eliminated the noise amplification and the aliasing artifacts. {In the row (\textbf{a}) of \cref{fig:seq}, the proposed method was demonstrated in prospective accelerations in parallel imaging and compressed sensing scenarios, confirmed the results in retrospective experiments. The performances of VN and the proposed methods were largely similar; however, the proposed method showed slight better preservation of boundaries between gray matter and white matter. More importantly, the proposed method uniquely supported different undersampling masks without the need for retraining the deep learning model.}
Table 1 presents the comparison of GRAPPA, {VN,} and the proposed method for knee (N = 170) and brain (N = 300) MRI testing images. With the increase of the undersampling factor, the PSNR of the proposed method decreased less, compared with that of GRAPPA. In addition, with acceleration factor R = 2 in brain MRI, the proposed method showed 8 dB more improvement in the PSNR than GRAPPA. In Supporting Information Video S2, the brain images of T2$^*$ weighting reconstructed with 3-fold prospective acceleration k-space data from a volunteer were presented.

\subsection{Compressed sensing reconstruction}
\begin{figure}[H]
	\centering
	\includegraphics[width=0.8\textwidth]{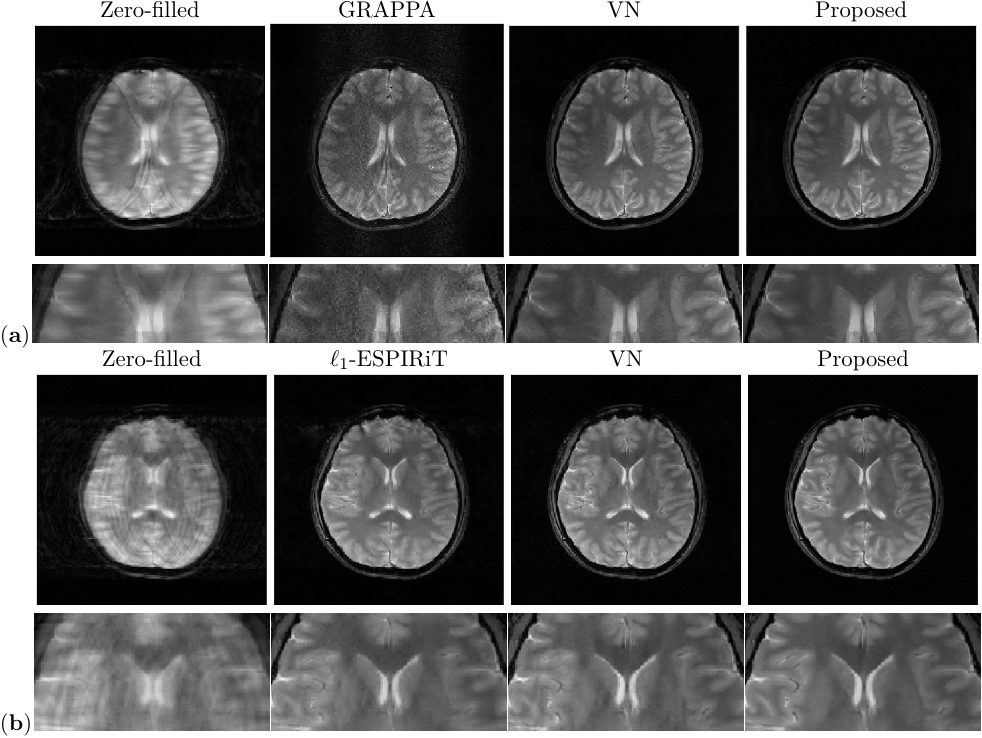}
	\caption{Comparison of different methods with prospective accelerations. The row (\textbf{a}) shows the images reconstructed from k-space data sampled in GRAPPA mask (R=3). The row (\textbf{b}) shows the images reconstructed from k-space data sampled in CS mask (R=22\%).}
	\label{fig:seq}
\end{figure}

\begin{figure}
	\centering
	\includegraphics[width=\textwidth]{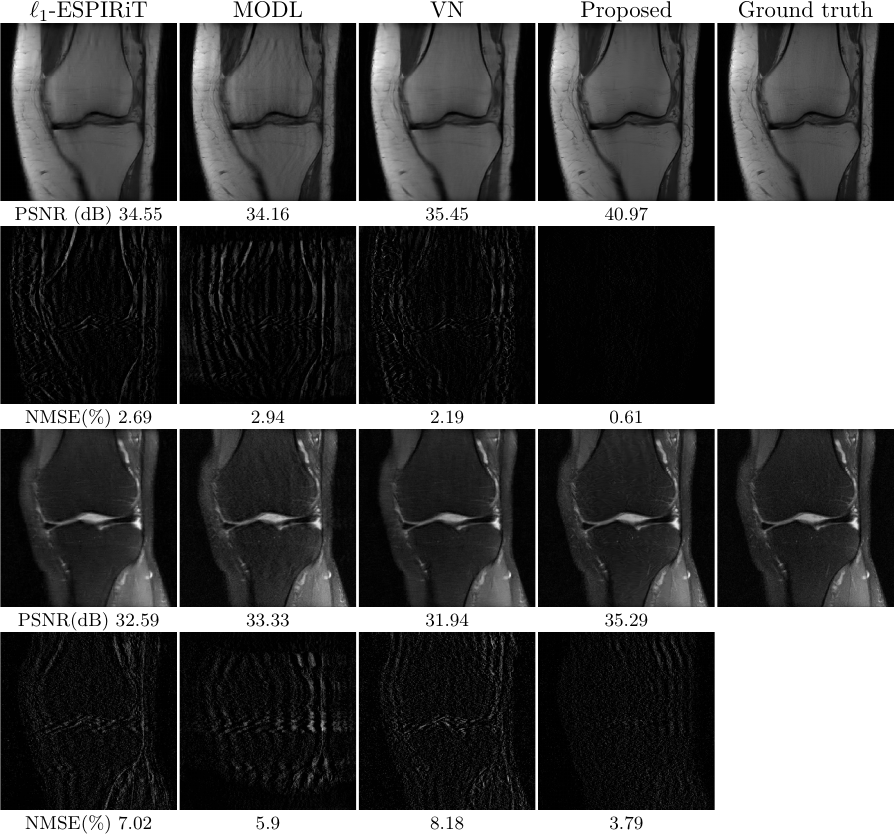}
	\caption{Comparison of different methods on PD and PDFS contrasts, using 27\% 1D undersampled k-space and $256\times256$ matrix size. The intensity of error maps was five times magnified. The proposed method substantially reduced the aliasing artifact and preserved image details in compressed sensing reconstruction.}
	\label{fig:cs_knee}
\end{figure}

\begin{figure}
	\centering
	\includegraphics[width=\textwidth]{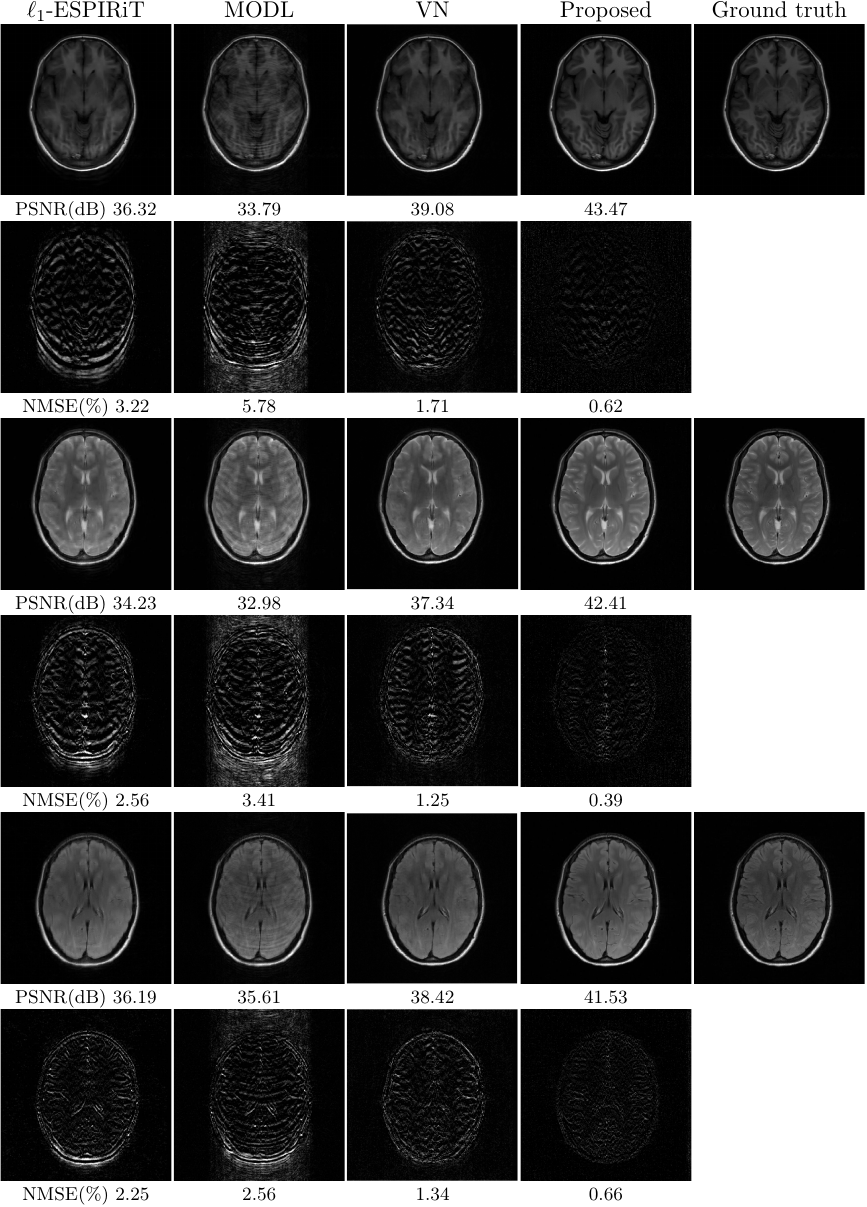}
	\caption{Comparison of compressed sensing and deep learning approaches for T1, T2, and FLAIR weighted image reconstructions, using 22\% 1D undersampled k-space and $256 \times256$ matrix size. The intensity of error maps was ten times magnified. The proposed method substantially reduced the aliasing artifact and preserved image details in compressed sensing reconstruction.}
	\label{fig:cs_brain}
\end{figure}

In \cref{fig:cs_knee} and \cref{fig:cs_brain}, the $\ell_1$-ESPIRiT and VN had caused apparent blurring in the reconstructed images for both knee and brain MRI data. Both the  $\ell_1$-ESPIRiT and MODL methods caused residual aliasing artifacts. Meanwhile, the proposed reconstruction recovered most anatomical structures and sharp boundaries in knee and brain MR images, compared with those from $\ell_1$-ESPIRiT, MODL {and VN} reconstructions, as shown on error maps in \cref{fig:cs_knee} and \cref{fig:cs_brain}. {Besides, the row (\textbf{b}) of \cref{fig:seq} shows that the proposed method had slight better edge preservation and artifact suppression compared with VN. This observation was consistent with our prospective results.}
Tables 1 summarized reconstruction results using $\ell_1$-ESPIRiT, MODL, {VN} and the proposed method. The proposed method generally showed more than 5 dB PSNR improvement compared with $\ell_1$-ESPIRiT, MODL and VN. In Supporting Information Video S3, the brain images of T2$^*$ weighting reconstructed with 22\% prospective k-space data from a volunteer were presented.

\begin{table}[H]
	\setlength{\tabcolsep}{0.4em}
	\centering
	\caption{PSNR/SSIM comparisons (in dB/\%, mean $\pm$ standard deviation) for parallel imaging, compressive sensing and the proposed method on knee and brain MRI.}
	\begin{tabular}{c c c c c c}
		\hline \hline
		\multicolumn{1}{l}{\textit{\small  Parallel imaging}}\\
		
		R factor & Organ & {GRAPPA}       & - & {VN}                  & {Proposed}\\ \hline
		\multicolumn{1}{l}{\textit{\small  PSNR}}\\
		R=2      & knee  & 35.24$\pm$4.53 & - & 38.23$\pm$2.35    &57.31$\pm$3.84 \\
		R=3      & knee  & 31.48$\pm$3.21 & - & 35.53$\pm$1.75    &45.64$\pm$2.20 \\
		R=4      & knee  & 26.39$\pm$1.67 & - & 33.52$\pm$1.37    &39.51$\pm$2.76 \\
		R=2      & brain & 38.05$\pm$4.73 & - & 46.19$\pm$3.03    &51.18$\pm$2.77 \\
		R=3      & brain & 31.60$\pm$3.33 & - & 43.67$\pm$2.74    &45.43$\pm$2.52 \\
		R=4      & brain & 28.27$\pm$2.84 & - & 40.51$\pm$2.39    &43.58$\pm$2.66 \\
		\multicolumn{1}{l}{\textit{\small  SSIM}}\\
		R=2      & knee  & 87.85$\pm$8.01 & - & 93.99$\pm$3.31     &99.65$\pm$0.32 \\
		R=3      & knee  & 76.82$\pm$10.56& - & 90.31$\pm$4.18     &97.95$\pm$1.18 \\
		R=4      & knee  & 62.34$\pm$6.47 & - & 86.39$\pm$4.33     &93.41$\pm$3.03 \\
		R=2      & brain & 83.48$\pm$3.00 & - & 97.90$\pm$1.37     &98.10$\pm$0.19 \\
		R=3      & brain & 69.24$\pm$5.53 & - & 96.74$\pm$2.03     &96.85$\pm$4.36 \\
		R=4      & brain & 58.32$\pm$6.56 & - & 95.31$\pm$2.45     &95.54$\pm$5.10 \\
		\hline
		\multicolumn{1}{l}{\textit{\small  Compressive sensing}}\\
		Sampling rate& Organ & {$\ell_1$-ESPIRiT} & {MODL}         & {VN} & Proposed\\ \hline 
		\multicolumn{1}{l}{\textit{\small PSNR}}\\
		15\% + 7\% & knee       & 30.81$\pm$2.01     & 26.78$\pm$3.19 & 31.87$\pm$0.92 & 35.34$\pm$3.13 \\
		20\% + 7\% & knee       & 31.81$\pm$2.23     & 31.30$\pm$2.93 & 30.72$\pm$0.62 & 37.14$\pm$3.23 \\
		15\% + 7\% & brain      & 32.73$\pm$3.46     & 29.06$\pm$3.24 & 33.95$\pm$2.25 & 39.80$\pm$2.73 \\
		20\% + 7\% & brain      & 34.51$\pm$3.94     & 30.70$\pm$3.25 & 34.73$\pm$2.26 & 41.18$\pm$2.70 \\
		\multicolumn{1}{l}{\textit{\small SSIM}}\\
		15\% + 7\% & knee       & 78.97$\pm$8.38  & 72.40$\pm$9.29 & 82.50$\pm$4.55  & 88.12$\pm$5.66 \\
		20\% + 7\% & knee       & 81.74$\pm$8.30  & 86.50$\pm$3.77 & 81.66$\pm$3.84  & 91.27$\pm$4.23 \\
		15\% + 7\% & brain      & 87.58$\pm$6.26  & 78.85$\pm$7.23 & 89.86$\pm$3.47  & 91.52$\pm$6.26 \\
		20\% + 7\% & brain      & 90.08$\pm$6.03  & 82.67$\pm$7.43 & 91.24$\pm$3.54  & 92.48$\pm$5.66 \\
		\hline 
	\end{tabular}
	\label{tab:all}
\end{table}

\subsection{Preliminary result in non-Cartesian MRI reconstruction}
\begin{figure}
	\centering
	\includegraphics[width=0.7\textwidth]{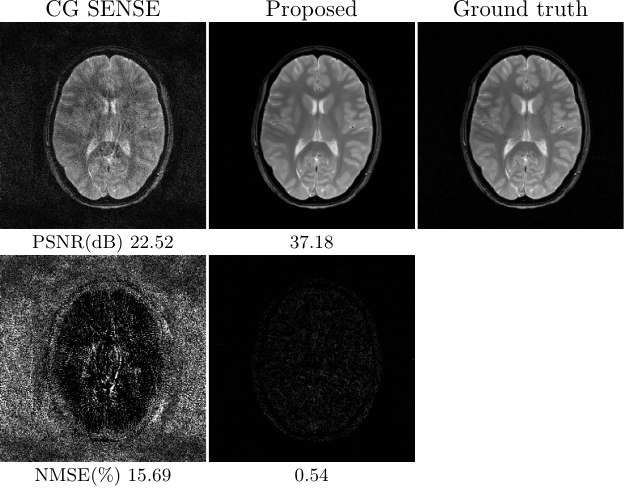}
	\caption{Comparison of the CG SENSE and proposed reconstruction for simulated spiral k-space with 4-fold acceleration (i.e., 6 out of 24 spiral interleaves), acquired by T2$^*$ weighted gradient echo sequence. The intensity of error maps was five times magnified. The proposed method substantially reduced the aliasing artifact in spiral reconstruction. Noted that the same deep learning model used in the previous Cartesian k-space reconstruction was applied to spiral reconstruction, without the need of re-training the deep learning model.}
	\label{fig:spiral}
\end{figure}

In this study, we used a T2$^*$ weighted gradient-echo images to simulate the spiral k-space data with 4-fold acceleration. The reconstructed images from the CG SENSE and the proposed method were compared in \cref{fig:spiral}. The proposed method showed apparent improvement regarding the aliasing artifact reduction and the preservation of T2$^*$ contrast between gray matter and white matter. The proposed method also showed a slight denoising effect on the reconstructed image compared with the ground truth.  Noted that the same deep learning model used in the previous Cartesian k-space reconstruction experiments in \cref{fig:pi_brain} and \cref{fig:cs_knee} was applied to spiral reconstruction, without the need of re-training the deep learning model.

\subsection{Phase maps}
\begin{figure}[H]
	\centering
	\includegraphics[width=0.75\textwidth]{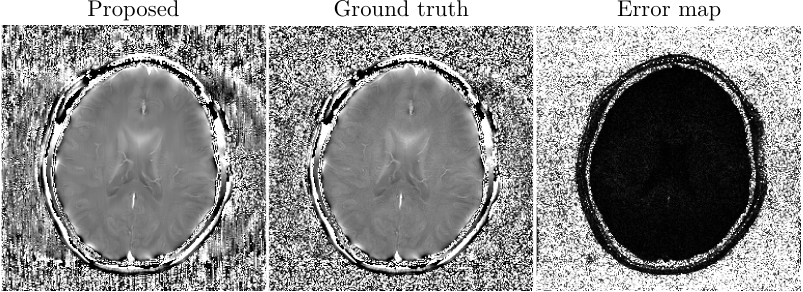}
	\caption{Phase maps from the ground truth and the proposed accelerated reconstruction with R = 4 and GRAPPA type of 1D undersampling. The raw images were acquired by T2$^*$ weighted gradient echo sequence. The proposed deep learning method preserved the major phase contrast even with high acceleration, R = 4. The error phase was 10 times magnified.}
	\label{fig:qsm}
\end{figure}

\cref{fig:qsm} shows the phase maps from the proposed accelerated reconstruction with 4-fold GRAPPA. Noted that the same deep learning model used in the previous brain experiments was applied to this experiment, with phase information preserved in all reconstructed images. The proposed deep learning method also showed {slight} de-noising effect on phase maps, while still preserved the major phase contrast even with high acceleration.
\section{Discussion}

The proposed method can reliably and consistently recover the nearly aliased-free images with relatively high acceleration factors. Meanwhile, as expected, the increase of image smoothing with high acceleration factors was noticed, reflecting the loss of intrinsic resolution. The estimated image from the maximum of the posterior can not guarantee the full recovery of the image details, i.e., PSNR $>$ 40 dB/SSIM$>$95\% for a full recovery. However, at modest acceleration, the reconstruction from a maximum of posterior showed the successful reconstruction of the detailed anatomical structures, such as vessels, cartilage, and membranes in-between muscle bundles. \par
In this study, the results demonstrated the successful reconstruction of high-resolution image (i.e., 256 $\times$ 256 matrix) with low-resolution prior (i.e., trained with 128 $\times$ 128 matrix), confirming the feasibility of reconstructing images of different sizes without the need for retraining the prior model. The prior model was trained by 128 x 128 images; it was still valid and applicable for the reconstruction of a high-resolution image. The proposed methods provided more than 8 dB improvement over the conventional GRAPPA reconstruction at the 4-fold acceleration in knee MRI. Besides, in contrast to other deep learning-based methods, which focused on the $\ell_2$ loss, the likelihood that was conditioned by pixel-wise dependencies of the whole image showed an improved representation capacity, leading to a higher reconstruction accuracy. The applicability of the proposed method in the patch-based reconstruction also suggested its high representation capacity and flexibility. Even when the inputs were image patches, the prior model could still recover the whole image.\par

\begin{figure}[H]
	\centering
	\includegraphics[width=0.7\textwidth]{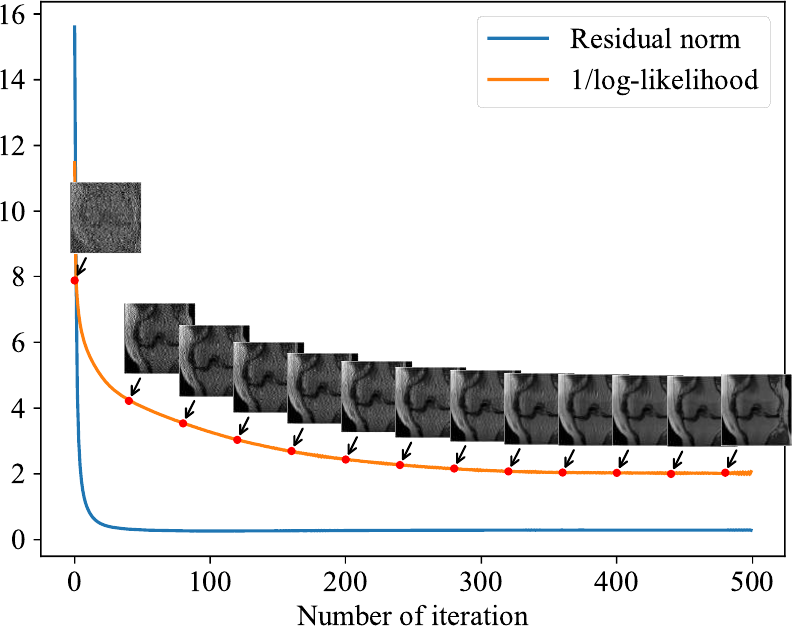}
	\caption{Convergence curves reflected stabilities of iterative steps: 1) maximizing the posterior, which effectively minimized the log-likelihood of MRI prior model and  2) k-space fidelity enforcement, which reduced the residual norm on k-space fidelity. The 22\% sampling rate and 1D undersampling scheme were used in this simulation. The residual norm was written as  $\|\boldsymbol{y}-\mathbf{A}\boldsymbol{x}\|_2^2$ in \cref{eq:13}, and the reciprocal of log-likelihood for MRI prior model was given in \cref{eq:g_prior}.} 
	\label{fig:curve}
\end{figure}

The projected subgradient approach to solving \cref{eq:12} was computationally inexpensive but converged slowly, as shown in \cref{fig:curve}. For a random initialization, the algorithm needed about 500 iterations to converge with a fixed step size. Meanwhile, we noticed that if the zero-filled-reconstructed image was used for initialization, the number of iterations required could be reduced to 100. Besides, the decay of residual norm stopped earlier than that of the log-likelihood, i.e., when the residual norm stopped decaying, the likelihood can still penalize the error. This evidence indicated that using the residual norm as the $\ell_2$ fidelity alone was sub-optimal, and the deep learning-based statistical regularization can lead to a better reconstruction result compared with the $\ell_2$ fidelity. Deep learning-based statistical regularization in the proposed method outperformed other conventional regularizations trained by image-level $\ell_2$ loss. $\ell_2$ loss did not give an explicit description of the relationship amid all pixels in the image, while the likelihood used in conjunction with the proposed image prior model was conditioned by the pixel-wise relationship and demonstrated superior performance compared with the conventional methods, under the current experimental setting.
Furthermore, the demonstrated image prior can be extended to a more elaborated form with clinical information, such as organs and contrast types, as the model inputs. For example, one could input the image prior with labels such as brain or knee. Then hypothetically, the image prior can be designed as a conditional probability for the given image label. In other words, the posterior would be dependent on both the k-space data and image labels. Moreover, the MR pulse sequence parameters could serve as image labels for the prior, such as echo time and repetition time. In short, the prior model can be used to describe clinical information or acquisition parameters. This setting opens up a future direction on a more elaborated image prior, incorporating clinical information and MR sequence parameters, for more intelligent image representation and pattern detection.\par

In this study, the generative network solely served as an image prior model, in contrast to how neural network was used in other deep learning-based reconstructions \cite{sun2016deep,schlemper2017deep,mardani2018deep,aggarwal2018model}. Specifically, in previous studies \cite{sun2016deep, schlemper2017deep,mardani2018deep,aggarwal2018model}, embedding k-space fidelity term into the network made the algorithm inflexible because image prior and undersampling artifacts were mixed during the training. The proposed method used the standard analytical term for fidelity enforcement; therefore, its flexibility was comparable to the traditional optimization algorithm, such as $\ell_1$ regularization. Due to unavoidable changes of the encoding scheme, e.g., the image size and the RF coils during MRI experiment in practice, it was essentially needed to separate the learned component (the image prior) from the encoding matrix used in the fidelity term in reconstruction. Besides, the proposed method showed the feasibility of incorporating the coil sensitivity information in the fidelity term, which enabled the changeable encoding scheme without the need of retraining the model \cite{pruessmann2001advances,fessler2010model}. In summary, the separation of the image prior and the encoding matrix embedded in the fidelity term made the proposed method more flexible and generalizable compared with conventional deep learning approaches.\par

\section{Conclusion}
In summary, this study presented the application of Bayesian inference in MR imaging reconstruction with the deep learning-based prior model. We demonstrated that the deep MRI prior model was a computationally tractable and effective tool for MR image reconstruction. The Bayesian inference significantly improved the reconstruction performance over that of conventional $\ell_1$ sparsity prior in compressed sensing. More importantly, the proposed reconstruction framework was generalizable for most reconstruction scenarios.

\section{Acknowledgment}
None

\bibliographystyle{unsrt}
\bibliography{ref}
\appendices
\section{Appendix A: derivation for the probability of logistic distribution in \cref{eq:4}}
The probability density function of logistic distribution is as follow
\begin{align}
f(x;\mu ,s)&={\frac {e^{-(x-\mu )/s}}{s\left(1+e^{-(x-\mu )/s}\right)^{2}}} \nonumber\\
&={\frac {1}{s\left(e^{(x-\mu )/(2s)}+e^{-(x-\mu )/(2s)}\right)^{2}}}\nonumber
\end{align}
The cumulative distribution function of logistic distribution is as follow
$$F(x;\mu ,s)=\int_{-\infty}^{x} f(x;\mu,s) = {\frac {1}{1+e^{-(x-\mu )/s}}}$$
Therefore if we assume the $\nu$ follows the logistic distribution, the probability of $\nu$ in range from $x-d/2$ to $x+d/2$ is
\begin{align}
P(\nu;\mu,s) & = \frac {1}{1+e^{-(\nu+d/2-\mu )/s}} - \frac {1}{1+e^{-(\nu-d/2-\mu )/s}}\nonumber\\
&=\sigma((\nu+d/2-\mu)/s) - \sigma((\nu-d/2-\mu)/s).\nonumber
\end{align}
Similarly, if we assume the $\nu$ follows the mixture logistic distribution
$$
\nu \sim \sum_{k=1}^{K} \pi_k \mathrm{logistic}(\mu_i, s_{k}),
$$
then, we have 
$$
P(\nu;\pi, \mu, s) = \sum_{k=1}^{K}\pi_k[\sigma((\nu+d/2-\mu_k)/s_k)-\sigma((\nu-d/2-\mu_k)/s_k)].
$$

\section{Appendix B: reconstruction for the varied image size with deep prior model}

\begin{algorithm}[H]
	\caption{Reconstruction algorithm with deep prior model}\label{alg:recon2}
	\hspace*{0.02in} \textbf{Input:} \\
	\hspace*{0.02in}	$\boldsymbol{y}$ - k-space data\\
	\hspace*{0.02in}	$\mathbf{A}$ - encoding matrix\\
	\hspace*{0.02in}  $\lambda$ - maximum iteration\\			
	\hspace*{0.02in} \textbf{Output:} \\
	\hspace*{0.02in} $\boldsymbol{x}$ - the restored image
	\begin{algorithmic}[1]
		\State Give a random initial point $\boldsymbol{x}^{(0)}, {k=1}$ \Comment{Initialization}
		\While{$\|g^{(k)}\|_2^2>\varepsilon$ and $k<\lambda$}\Comment{Iteration}
		\State Generate a random shifting offset $d$
		\State Shift $\boldsymbol{x}^{(k)}$ $d$ pixels away from the center circularly
		\State Split $\boldsymbol{x}^{(k)}$ into pieces $\boldsymbol{s}^{(k)}$ for feeding to network
		\State Get subgradient  $\nabla_{\boldsymbol{x}}\log(g(\boldsymbol{x}))$ at $\boldsymbol{x}^{(k)}$
		\State Pick a step size $\alpha_k=1/k$ {or use a fixed step size}
		\State Update $\boldsymbol{s}^{(k+1)}=\boldsymbol{s}^{(k)}-\alpha_k \nabla_{\boldsymbol{s}^{(k)}}\log{g(\boldsymbol{s}^{(k)})}$
        \State Merge pieces $\boldsymbol{s}^{(k+1)}$ into $\boldsymbol{z}^{(k+1)}$ for projection
        \State Shift $\boldsymbol{z}^{(k+1)}$ $-d$ pixels away from the center circularly
		\State Projection $\boldsymbol{x}^{(k+1)}={\underset {\boldsymbol{x}\in X }{\operatorname {arg\,min} }}\frac{1}{2}\|\boldsymbol{x}-\boldsymbol{z}^{(k+1)}\|_2^2$
		\State {$k=k+1$}
		\EndWhile\label{euclidendwhile}
		\State \textbf{return} $\boldsymbol{x}^{(k+1)}$
	\end{algorithmic}
\end{algorithm}

\section{List of supporting information}
The following support information will be available on this webpage\footnote{\url{https://github.com/mrirecon/spreco}}$^,$\footnote{\url{https://onlinelibrary.wiley.com/doi/full/10.1002/mrm.28274}}.

\begin{itemize}
	\item Supporting Information Table S1: The scan parameters of different weightings used in brain MRI experiments.
	\item Supporting Information Figure S1: The diagram shows the PixelCNN++ network in [13], which was the prior model used in this study, i.e., $g(\boldsymbol{x})$ in  \cref{eq:g_prior}. Each ResNet block (gray) has a gated-resnet components. The last layer is fully connected. The input of network was $\boldsymbol{x}$, outputs of network were parameters of mixture distribution $(\boldsymbol{\pi, \mu, s, \alpha})$, which were fed into the conditional probability model in \cref{eq:p_model}
\item Supporting Information Figure S2: k-space masks used in the compressed sensing, parallel imaging, and deep learning reconstructions. Bright lines indicate the sampled frequency encoding lines in the 2D k-space, i.e., the 1D undersamplings were simulated.
\item Supporting Information Video S1: The evolution of image during the process of reconstruction.
\item Supporting Information Video S2: Reconstruction by the proposed method of a healthy volunteer for T$_2^*$-weighted brain images with undersampling factor R=3 (GRAPPA prospective experiment).
\item Supporting Information Video S3: Reconstruction by the proposed method of a healthy volunteer for T$_2^*$-weighted brain images with 22\% k-space (Compressie sensing prospective experiment).
\end{itemize}

\iftoggle{SUPMATERIAL}
{
\includepdf[pages=-]{Supplementary/Suplementary_Material.pdf}
}

\end{document}